\documentclass[letterpaper, 10 pt, conference]{ieeeconf}  

\IEEEoverridecommandlockouts                              

\usepackage{graphics} 
\usepackage{epsfig} 
\usepackage{mathptmx} 
\usepackage{times} 
\usepackage{amsmath,amsfonts}
\usepackage{amssymb}  

\usepackage{url}
\usepackage{booktabs}
\usepackage{siunitx}
\usepackage{csquotes}
\usepackage{pgf}
\usepackage{pgfplots}
\usepackage[T1]{fontenc}
\usepackage{tikz}
\usepackage{makecell}
\usepackage{kotex} 
\usepackage{multirow} 
\usepackage{pifont} 
\usepackage{arydshln} 
\usepackage{subfiles}
\usepackage{balance} 
\usepackage[absolute]{textpos} 
\usepackage{xcolor}
\usepackage{dsfont}
\usepackage{adjustbox}  

\usepackage[caption=false, font=footnotesize]{subfig}

\let\labelindent\relax
\usepackage{enumitem}

\title{\LARGE \bf
Technical Report for ICRA 2025 GOOSE 2D Semantic Segmentation
Challenge: Boosting Off-Road Segmentation via Photometric Distortion and Exponential Moving Average
}

\author{Wonjune Kim, Lae-Kyoung Lee and Su-Yong An
\thanks{This work was supported by Electronics and Telecommunications Research Institute (ETRI) grant funded by the Korean government [25ZD1160, Development of ICT Convergence Technology for Daegu-Gyeongbuk Regional Industry]. \newline
\hspace*{1em}The authors are with Daegu-Gyeongbuk Research Center, Electronics and
Telecommunications Research Institute (ETRI), Daegu 42994, South Korea (email: wonjune.kim@etri.re.kr; laeklee@etri.re.kr; syong.an@etri.re.kr). 
}}

\usepackage{fancyhdr}
\IEEEoverridecommandlockouts      

\newcommand{\goosefooter}{%
  \footnotesize Winners of the GOOSE 2D Semantic Segmentation Challenge at the IEEE ICRA Workshop on Field Robotics 2025}

\fancypagestyle{withfooter}{%
  \fancyhf{}

  \fancyfoot[C]{\goosefooter}
}

\fancypagestyle{plain}{%
  \fancyhf{}

  \fancyfoot[C]{\goosefooter}
}

\begin{document}

\maketitle
\thispagestyle{withfooter}
\pagestyle{withfooter}

\begin{abstract}

We report on the application of a high-capacity semantic segmentation pipeline to the GOOSE 2D Semantic Segmentation Challenge for unstructured off-road environments. Using a FlashInternImage-B backbone together with a UPerNet decoder, we adapt established techniques, rather than designing new ones, to the distinctive conditions of off-road scenes. Our training recipe couples strong photometric distortion augmentation (to emulate the wide lighting variations of outdoor terrain) with an Exponential Moving Average (EMA) of weights for better generalization. Using only the GOOSE training dataset, we achieve 88.8\% mIoU on the validation set.

\end{abstract}

\section{Introduction}
Autonomous navigation in off-road environments requires a perception system that can accurately delineate traversable terrain, vegetation, and obstacles under extreme and rapidly changing weather and lighting.
Compared with urban scenes, off-road imagery exhibits
greater appearance diversity (e.g.\ mud, snow, dense underbrush) and fewer structural cues (absence of lane markings or curbs), making
pixel-level interpretation markedly harder.

To enable benchmarking in this domain, the GOOSE \cite{GOOSE-Dataset, GOOSE-Ex} dataset provides seasonally diverse RGB frames, each annotated with a
fine-grained 64-class label map. For the ICRA~2025 GOOSE 2D Semantic-Segmentation Challenge these labels are consolidated into nine operational categories (\textit{vegetation}, \textit{natural ground}, \textit{artificial ground}, \textit{artificial structures}, \textit{obstacle}, \textit{vehicle}, \textit{human}, \textit{sky}, \textit{other}).

Two factors make the task especially challenging:
\begin{itemize}
  \item \textbf{Severe class imbalance:} Approximately \(90\%\) of the pixels belong to only three classes \textit{vegetation}, \textit{terrain} and \textit{sky}, while safety-critical but visually small objects (\textit{obstacle}, \textit{human}) are under-represented.
  \item \textbf{Ambiguous, low-contrast boundaries:} Natural materials often blend gradually (e.g.\ grass–soil, water–mud), confounding edge-based segmentation cues and reducing the effectiveness of standard cross-entropy optimization.
\end{itemize}

We combine established components that, when carefully tuned, prove
effective in the off‑road domain.
The backbone is FlashInternImage‑B \cite{flashinternimage}, obtained by augmenting InternImage‑B \cite{wang2023internimage} with DCNv4 \cite{flashinternimage} deformable convolutions, and it paired with a multi‑scale UPerNet \cite{upernet} decoder.
Training employs \(2048 \times 2048\) crops drawn with scale jitter; color robustness is enhanced through photometric distortion, and an exponential moving average of the parameters improves stability in the presence of label noise.
The experimental analysis in Sec.~\ref{sec:experiments} confirms that this configuration achieves competitive performance on the GOOSE 2D benchmark, particularly for classes that are poorly represented in the training data such as \textit{obstacle} and \textit{human}.

\section{Method}
\label{sec:method}

\begin{table*}[t!]
	\vspace{3pt}
	\centering
	\caption{Per-class and mean IoU (mIoU) on the GOOSE 2D Semantic Segmentation Challenge validation set. Starting from a baseline FlashInternImage-B model, we successively add Photometric Distortion and Exponential Moving Average (EMA).}
	\renewcommand{\arraystretch}{1.1}
	\begin{tabular}{*{8}{cllcccccccccc}}
		\toprule
		                                              & network & \textbf{mIoU}$\,\uparrow$ & Other & \makecell{Artificial \\ Structure} & \makecell{Artificial \\ Ground} & \makecell{Natural \\ Ground} & Obstacle & Vehicle & Vegetation & Human & Sky \\
		\midrule
		\multirow{4}{*} &
		{FlashInternImage-B~\cite{flashinternimage}}
		                                              & 87.28
		                                              & 91.18
		                                              & 79.31
		                                              & 93.6
		                                              & 89.34
		                                              & 76.18
		                                              & 89.73
		                                              & 88.35
		                                              & 80.32
		                                              & 97.47
		\\
          & {+ Photometric distortion}
		                                              & 87.76
		                                              & 92.04
		                                              & 80.37
		                                              & 92.48
		                                              & 89.08
		                                              & 76.89
		                                              & 90.71
		                                              & 88.88
		                                              & 81.89
		                                              & 97.53
            \\
          & {+ Photometric distortion + EMA}
		                                              & \textbf{88.88}
		                                              & \textbf{93.62}
		                                              & \textbf{81.61}
		                                              & \textbf{94.29}
		                                              & \textbf{89.6}
		                                              & \textbf{78.68}
		                                              & \textbf{91.78}
		                                              & \textbf{88.89}
		                                            & \textbf{83.83}
		                                              & \textbf{97.63}             
		\\
		\bottomrule
	\end{tabular}
	\label{tab:evaluation}
\end{table*}

\begin{figure*}[!t]            
  \centering
  \includegraphics[width=\textwidth]{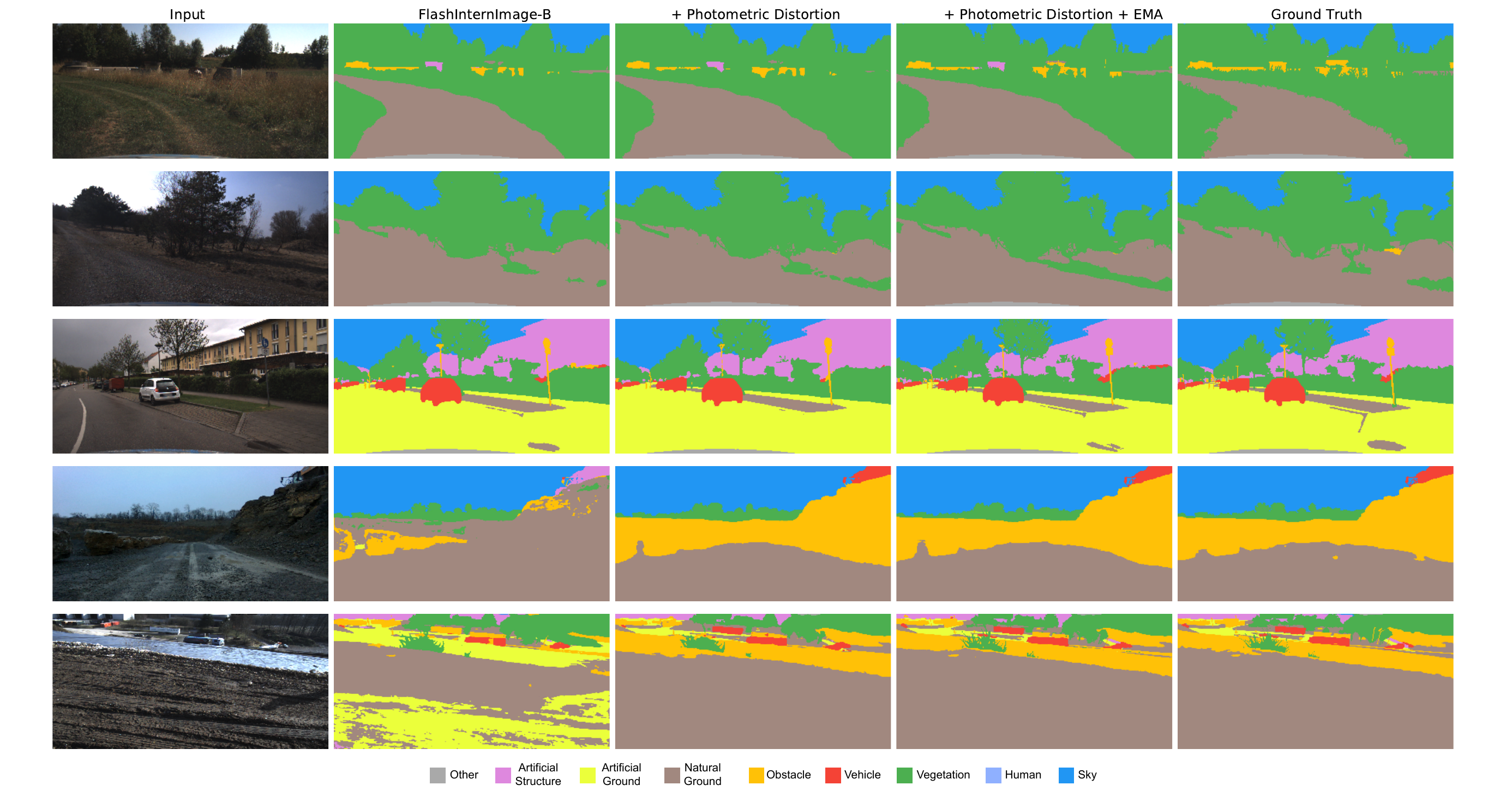}
  \caption{\textbf{Qualitative comparison on the validation set.}  
Columns from left to right: (1) input RGB image, (2) prediction of the FlashInternImage-B baseline, (3) baseline plus photometric distortion, (4) baseline plus photometric distortion and EMA, and (5) ground-truth
annotation.}
  \label{fig:qualitative}
\end{figure*}

\subsection{Baseline}
\label{sec:baseline}
The proposed method adopts a \emph{FlashInternImage-B} backbone, in which every deformable-convolution layer of InternImage-B is upgraded from DCNv3 \cite{wang2023internimage} to the faster DCNv4 \cite{flashinternimage} operator.  Thanks to this replacement, each training iteration is approximately \(1.8\times\) faster while accuracy is preserved.
Feature maps at \(\tfrac14\), \(\tfrac18\), \(\tfrac1{16}\), and \(\tfrac1{32}\) of the input
resolution are aggregated by the \emph{UPerNet} \cite{upernet} decoder, whose FPN branch merges
multi-scale information and whose PSP branch captures global context.
The head produces nine logits, one per GOOSE class, which are bilinearly upsampled to the full image size.
Pixel-wise soft-max cross-entropy is used as the optimization target.

Training is performed with \texttt{AdamW} (initial
\(\text{lr}=6\!\times\!10^{-5}\)) under a poly learning-rate schedule for
96k iterations (\(\approx 150\) epochs).
The Images are randomly scaled in the range \([0.5,2.0]\) and then
cropped or padded to \(2048\times2048\).

\subsection{Photometric Distortion}
\label{sec:photometric}
\begin{figure}[!t]            
  \centering
  \includegraphics[width=\columnwidth]{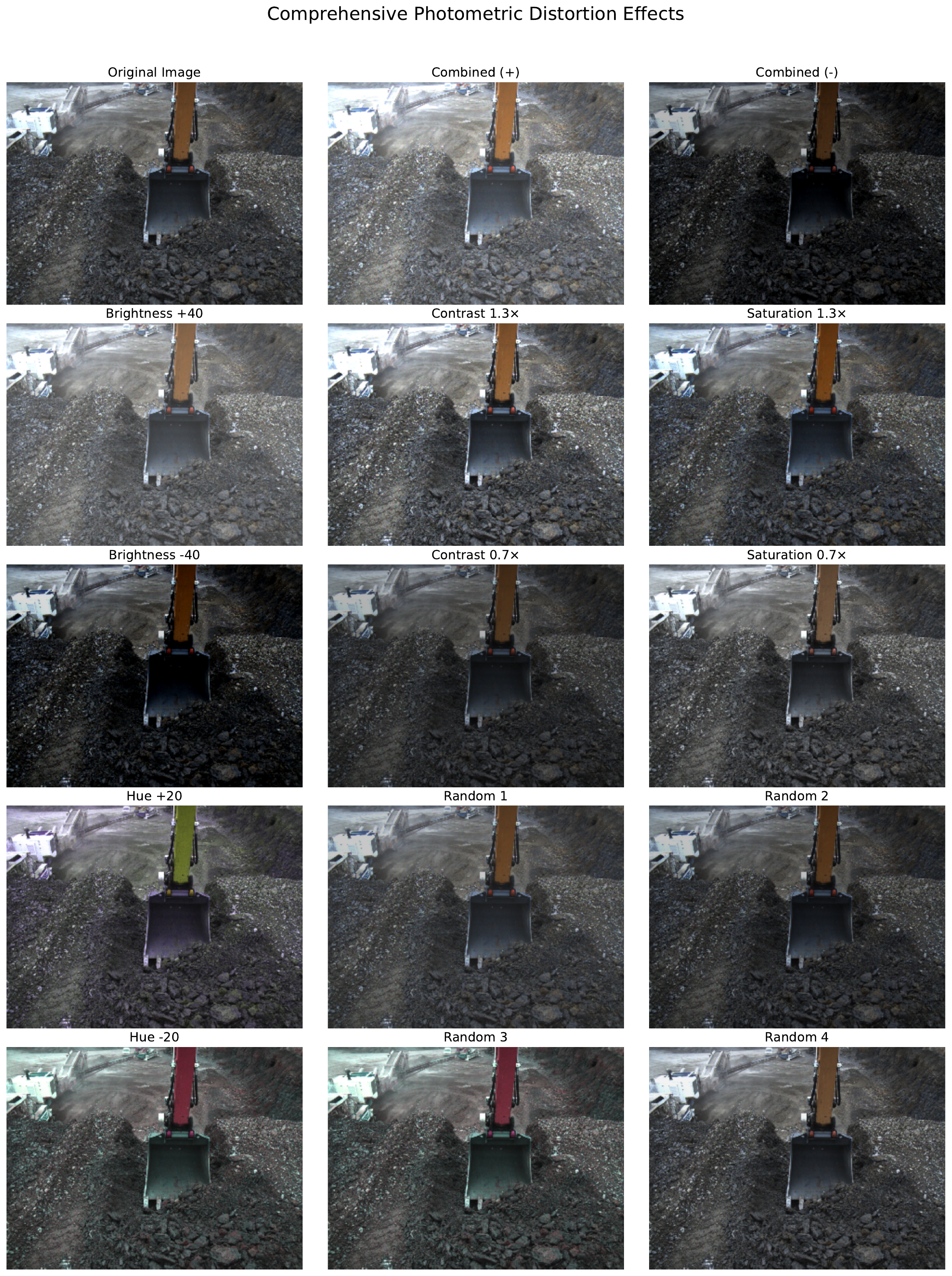}
  \caption{\textbf{Comprehensive Photometric Distortion Effects.} From left to right and top to bottom the grid shows (i) the original RGB image, (ii) a \emph{combined (+)} sample where brightness, contrast, saturation and hue are jointly increased,
(iii) a \emph{combined (–)} sample where the same factors are jointly decreased,
followed by isolated adjustments of
brightness ($+40$ and $-40$),
contrast ($\times1.3$ and $\times0.7$), and
saturation ($\times1.3$ and $\times0.7$).
These transformations are drawn at random during training, each with probability 0.5, to expose the network to the full range of illumination and color conditions encountered in off-road scenes.}  
  \label{fig:photometric}
\end{figure}

GOOSE~\cite{GOOSE-Dataset, GOOSE-Ex} images exhibit large variations in illumination, ranging from dark forest scenes to bright snow fields, which produce notable color shifts.
To enhance robustness we apply \texttt{PhotoMetricDistortion} during
training: brightness, contrast, saturation, and hue are each perturbed
independently with probability~\(0.5\) using uniformly sampled factors
(see Fig.~\ref{fig:photometric} for examples).
These stochastic color transforms broaden the appearance
distribution that the network observes, encouraging it to rely on shape
and texture rather than raw color cues.
Compared with purely geometric augmentation, the additional photometric
jitter raises validation performance by \(+0.48\) mIoU.

\subsection{Exponential Moving Average}
\label{sec:ema}
To stabilize optimization and mitigate label noise we maintain an
\emph{Exponential Moving Average} (EMA) of the network parameters,
updated each iteration as
\[
  \boldsymbol{\theta}_{\textsc{ema}}^{(t)} \;=\;
  \alpha\,\boldsymbol{\theta}_{\textsc{ema}}^{(t-1)} +
  (1-\alpha)\,\boldsymbol{\theta}_{\textsc{current}}^{(t)},
  \qquad \alpha = 0.999 .
\]
The EMA snapshot is used for every validation check-point and for the
final evaluation.
When applied on top of the photometric distortion baseline, EMA brings an additional \(\,+1.08\) mIoU and visibly suppresses speckle
artifacts in large homogeneous regions.

\section{Experiments}
\label{sec:experiments}
\subsection{Dataset and Training Protocol}
For all trials we merge the original \textsc{GOOSE} training split
(\(\approx 8\,\mathrm{k}\) images) with the recently released \textsc{GOOSE-EX}
training split (\(\approx 4\,\mathrm{k}\) images), then hold out the official \textsc{GOOSE} and \textsc{GOOSE-EX} validation (\(\approx 1.4\,\mathrm{k}\) images) for evaluation.
Models are trained on four NVIDIA RTX 3090 GPUs
(batch\_size\,=\,2 per GPU, mixed precision) for
\(96\,\mathrm{k}\) iterations with the optimization recipe described in Sec.~\ref{sec:method}.  Performance is reported as the mean Intersection-over-Union (mIoU) averaged across the nine challenge
classes.

\subsection{Quantitative Results}

Table~\ref{tab:evaluation} summarizes the effect of each training refinement.
Starting from the FlashInternImage-B \cite{flashinternimage} baseline, photometric distortion adds
\(0.48\) mIoU. Subsequent application of EMA yields a further
\(1.12\) mIoU, giving a total improvement of \(1.60\) points on the validation
set.  Per-class scores reveal that
\textit{obstacle} and \textit{human} benefit the most from EMA, while
photometric jitter chiefly enhances \textit{sky} and \textit{other}.
\subsection{Qualitative Results}
Rows 1 and 2 of Fig. \ref{fig:qualitative} are dominated by the frequent classes natural ground and vegetation; as these classes account for most pixels, all three models produce nearly identical, accurate masks.
In Row 4 the baseline mislabels the rock–pile on the right-hand side of the road as natural ground, whereas the variant trained with photometric distortion correctly assigns it to the obstacle class.
Row 5 further highlights the benefit of the augmentations: the baseline confuses artificial ground with natural ground, while the photometric-distortion model separates the two classes cleanly; adding EMA sharpens the class boundaries even more, yielding the crispest segmentation among the three variants.

\section{Conclusions}
We have presented a practical yet high-performing off-road semantic-segmentation method that combines a FlashInternImage-B backbone, a multi-scale UPerNet
decoder, photometric-distortion augmentation and an exponential moving average of the weights.
Ablation studies show that photometric distortion and EMA contribute \(0.48\) mIoU and \(1.12\) mIoU respectively on the merged GOOSE, GOOSE-EX validation split, yielding a final
score of 88.8 mIoU (Table~\ref{tab:evaluation}).
Qualitative results (Fig.~\ref{fig:qualitative}) confirm sharper boundaries
and more reliable predictions for under-represented classes such as
\textit{obstacle} and \textit{human}.

When submitted to the official GOOSE 2D Challenge evaluation server our model attains \textbf{84.5 mIoU} on the test set of the ICRA~2025 \textit{GOOSE 2D Semantic-Segmentation Challenge}, placing \textbf{second} on the public leaderboard.




\bibliographystyle{IEEEtran}
\bibliography{root}


\end{document}